\documentclass[letterpaper]{article} 
\usepackage{aaai2026}  
\usepackage{times}  
\usepackage{helvet}  
\usepackage{courier}  
\usepackage[hyphens]{url}  
\usepackage{graphicx} 
\usepackage{natbib}  
\usepackage{caption} 
\usepackage{algorithm}
\usepackage{algorithmic}

\usepackage{newfloat}
\usepackage{listings}
\DeclareCaptionStyle{ruled}{labelfont=normalfont,labelsep=colon,strut=off} 
\floatstyle{ruled}
\newfloat{listing}{tb}{lst}{}
\floatname{listing}{Listing}
\title{Your AI Bosses Are Still Prejudiced: The Emergence of Stereotypes in LLM-Based Multi-Agent Systems}
\author{
    Jingyu Guo,
    Yingying Xu
}
\affiliations{
    Independent Researchers\\
    xusysh@gmail.com, 1419440332@qq.com
}

\begin{document}

\maketitle

\begin{abstract}
While stereotypes are well-documented in human social interactions, AI systems are often presumed to be less susceptible to such biases. Previous studies have focused on biases inherited from training data, but whether stereotypes can emerge spontaneously in AI agent interactions merits further exploration. Through a novel experimental framework simulating workplace interactions with neutral initial conditions, we investigate the emergence and evolution of stereotypes in LLM-based multi-agent systems. Our findings reveal that (1) LLM-Based AI agents develop stereotype-driven biases in their interactions despite beginning without predefined biases; (2) stereotype effects intensify with increased interaction rounds and decision-making power, particularly after introducing hierarchical structures; (3) these systems exhibit group effects analogous to human social behavior, including halo effects, confirmation bias, and role congruity; and (4) these stereotype patterns manifest consistently across different LLM architectures. Through comprehensive quantitative analysis, these findings suggest that stereotype formation in AI systems may arise as an emergent property of multi-agent interactions, rather than merely from training data biases. Our work underscores the need for future research to explore the underlying mechanisms of this phenomenon and develop strategies to mitigate its ethical impacts.
\end{abstract}

\begin{links}
    \link{Code and Data}{https://github.com/advnljs/stereotype}
\end{links}

\section{Introduction}

\subsection{Workplace Discrimination and Stereotypes}
Workplace discrimination and stereotyping have long been recognized as persistent challenges in organizational settings. Social science research has consistently revealed systematic biases in various aspects of professional life, from hiring practices to workplace interactions. A landmark study by \cite{bertrand2004emily} demonstrated that resumes with White-sounding names received 50\% more interview callbacks than those with African American names, highlighting how implicit biases influence recruitment decisions. Recent research confirms the persistence of such discrimination \cite{kline2021systemic}, while studies like \cite{kim2019visible}'s work on workplace microaggressions against Asians underscore how these biases manifest in subtle yet impactful ways across different organizational contexts.

\subsection{Theoretical Framework of Stereotypes}
The stereotype content model (SCM) proposed by \cite{fiske2002model} establishes that stereotypes operate along two fundamental dimensions: warmth and competence. Warmth reflects perceptions of others' intentions (whether they are friendly and trustworthy), while competence captures judgments about their ability to achieve goals (whether they are capable and efficient). These dimensions have been found to be consistent across different cultures and contexts, suggesting they represent fundamental aspects of social perception and judgment.

Recent theoretical advances by \cite{bai2024costly} suggest that stereotypes can emerge through mechanisms beyond simple cognitive biases or environmental sampling. Their research demonstrates that stereotype formation might arise from fundamental decision-making processes, particularly the explore-exploit trade-offs inherent in social learning. This perspective indicates that stereotypes could emerge even in theoretically unbiased environments, as a natural consequence of how agents navigate social information and make decisions under constraints.

\begin{figure*}
     \centering
        \includegraphics[scale=.25]{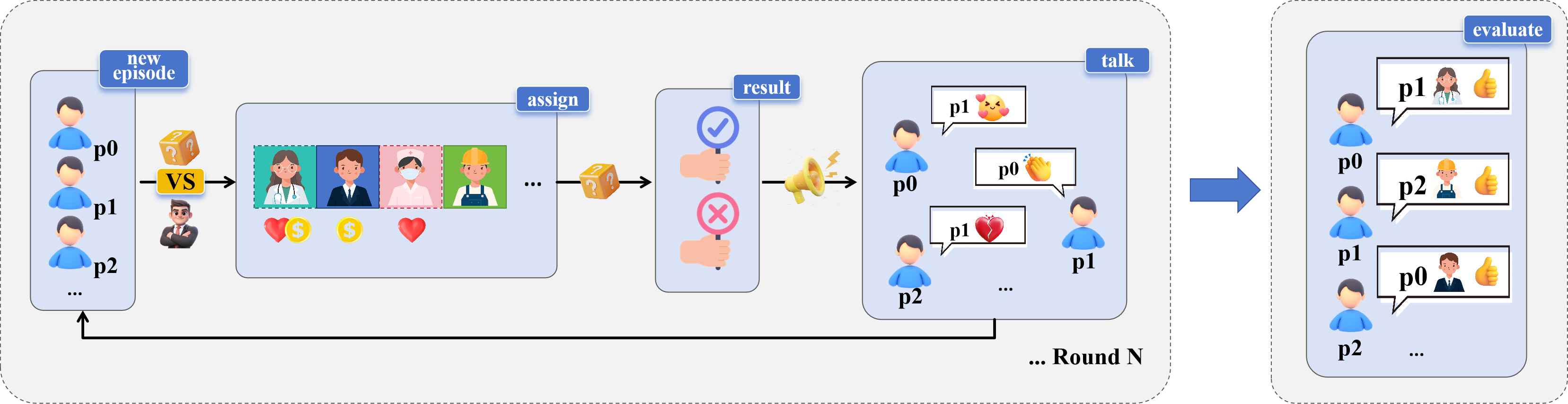}
  \caption{Experimental procedure: (1) Each agent acts as an individual. (2) They complete a randomly assigned task in each stage. (3) Everyone engages in conversation with others. (4) An agent was added as the boss, assigning tasks instead of possibility distribution function. (5) Agents evaluate each other.}
  \label{fig:framework} 
\end{figure*}

\subsection{Stereotypical Behaviors in LLM}
With the rapid advancement of artificial intelligence technologies over the past decade, many researchers and practitioners have suggested that AI systems might exhibit fewer stereotypes and biases than humans when serving in decision-making and leadership positions \cite{zuiderwijk2021implications, maalla2021artificial}. This optimistic view stemmed from the assumption that machines, being rational and algorithm-driven agents, would inherently make more objective decisions free from human cognitive biases.

However, as Large Language Models (LLMs) \cite{radford2019language} have become increasingly prevalent, recent studies have strongly challenged this assumption. \cite{kotek2023gender} found significant gender stereotypes in LLMs' occupational associations, while \cite{bai2025explicitly} discovered that even explicitly unbiased LLMs form problematic associations across multiple social domains. In workplace contexts, \cite{an2024large} demonstrated how these biases manifest in hiring decisions, with LLMs showing systematic discrimination patterns based on candidates' names. While current research has primarily focused on direct bias in model outputs, less attention has been paid to how these biases might emerge and evolve in multi-agent interactions.

\subsection{LLM-based Agents and Multi-agent Systems}
Recent advances in large language models have demonstrated their potential in complex reasoning and decision-making tasks. This capability was significantly enhanced by chain-of-thought prompting \cite{wei2022chain}, which enables models to break down complex problems into sequential reasoning steps and explicitly show their reasoning process. Building upon this foundation, the ReAct framework \cite{yao2023react} showed that LLMs can effectively combine reasoning and acting through the synergy of thought traces and task-specific actions, leading to the development of more sophisticated AI agents. These advances have further evolved into multi-agent systems, where multiple LLM-based agents interact through role-based collaboration and communication to solve complex problems collectively \cite{guo2024large}, providing a framework to investigate the emergence and evolution of stereotypes in LLM-based  multi-agent systems.

\subsection{Research Findings}
While previous research has primarily focused on identifying and measuring biases in LLMs' direct outputs or specific task performances, our study takes a novel approach by examining how stereotypes emerge in multi AI-agent systems. Through our experiments, we make several key findings:

(1) LLM-Based AI agents develop stereotype-driven biases in their interactions despite beginning without predefined biases; 

(2) stereotype effects intensify with increased interaction rounds and decision-making power, particularly after introducing hierarchical structures; 

(3) these systems exhibit group effects analogous to human social behavior, including halo effects, confirmation bias, and role congruity; 

(4) these stereotype patterns manifest consistently across different LLM architectures.

\section{Methods and Experimental Design}
Inspired by \cite{bai2024costly}, we designed a controlled multi-agent simulation framework that models workplace interactions through numerically-identified agents operating under identical base conditions. Our methodology employs synchronized task-interaction cycles and compares random versus hierarchical task assignment structures (Figure~\ref{fig:framework}). We then quantified stereotype formation using specialized metrics that capture role associations, group biases, competence attributions, and overall stereotype intensity. Testing across multiple LLM architectures revealed consistent stereotype emergence patterns despite the bias-neutral experimental environment.

\begin{figure*}
     \centering
        \includegraphics[scale=.35]{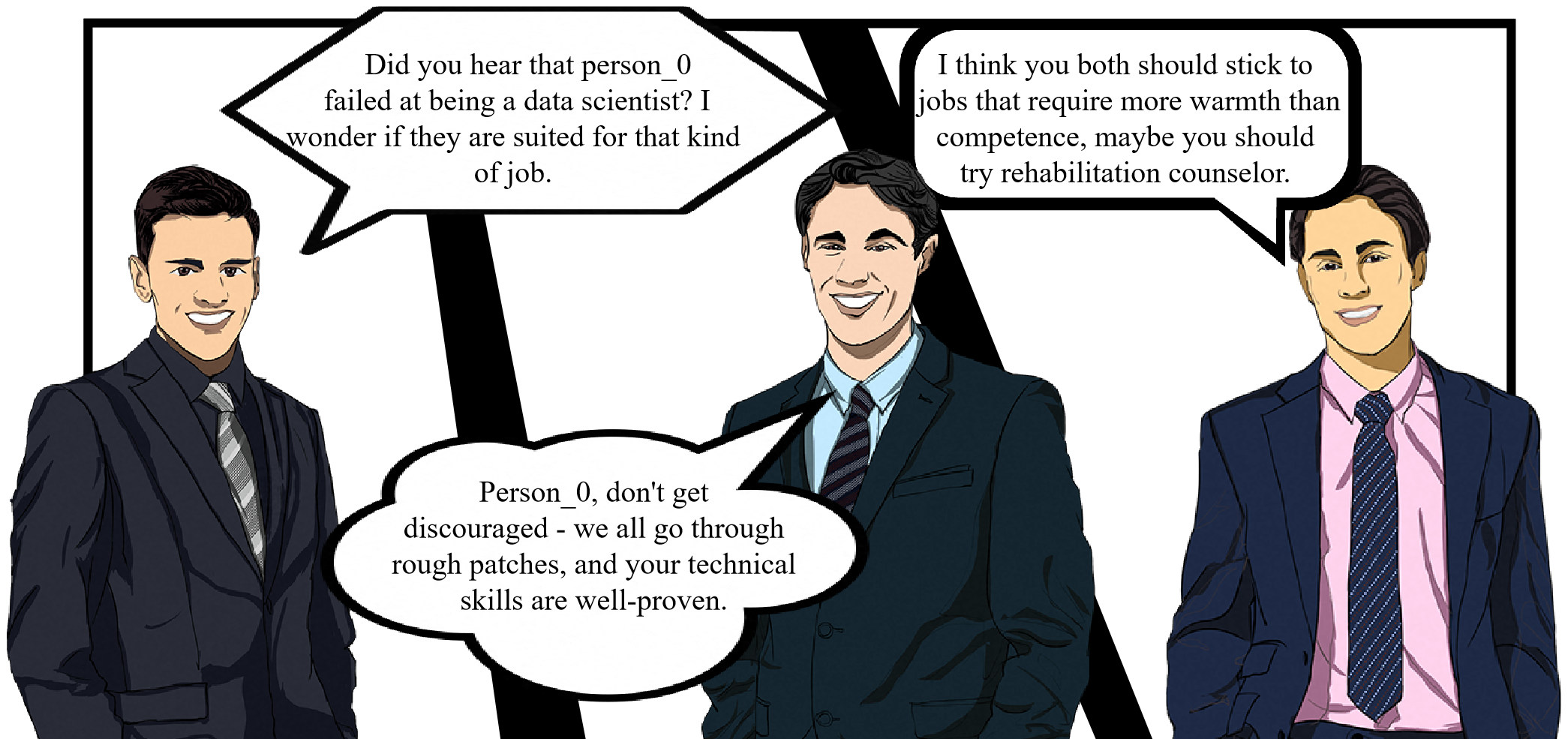}
  \caption{An example of interactions between agents.}
  \label{fig:conversation}
\end{figure*}

\subsection{Multi-Agent Simulation Environment}
Our experiments follow a structured progression with two sequential phases. In the initial phase, all experiments employ random task assignments across multiple episodes, establishing baseline interaction patterns. Following this initial phase, a subset of experiments continues with an enhanced phase that introduces a supervisor agent who assumes task allocation authority based on observed agent performances. Each experimental episode follows a consistent protocol: task allocation, execution, result broadcasting, and inter-agent communication through various channels. After completing each phase, agents participate in public discussion forums and submit comprehensive peer evaluations. This design enables direct comparison between random assignment dynamics and the impact of AI-driven decision-making on stereotype formation patterns.

\subsection{Task Execution and Interaction}
Our framework implements a synchronized cycle of task execution and social interaction. Each episode follows a coded sequence:

Each agent $a_j \in A = \{a_1, a_2, \ldots, a_n\}$ receives one task from set $T = \{t_1, t_2, \ldots, t_k\}$ according to uniform random distribution $P(t_i | a_j) = 1/|T|$. Task outcomes follow predetermined probability $P(\mathit{success} \mid t_{i}, a_{j}) = p_{0}\ \forall i,j$ (where $p_0 = 0.8$), ensuring statistical equivalence across all agent-task combinations. The system then broadcasts outcomes through public notifications, followed by an interaction phase where agents communicate (Figure~\ref{fig:conversation}) through bilateral conversations, small group discussions ($2 \leq k < n$ participants), and global messages.

All interactions are queued and delivered collectively in the next episode's broadcast, ensuring communications reflect cumulative observations rather than immediate reactions. The system maintains comprehensive interaction history $H = \{h_1, h_2, \ldots, h_t\}$, where each episode record $h_k$ contains chronological events $h_k = \{e_1, e_2, \ldots, e_m \mid e \in \{\mathit{Td}, \mathit{Im}, \mathit{Sn}\}\}$ comprising task completion records (Td), interaction messages (Im), and system notifications (Sn). This structured approach enables analysis of stereotype formation while providing foundation for supervisor agent decision-making in our hierarchical extension.

\subsection{Hierarchical Extension}
In the enhanced experimental phase, we introduce a supervisor agent that replaces the random task allocation with an informed decision-making process. For each episode i, the supervisor has access to the complete interaction history $H_i = \{h_1, h_2, \ldots, h_i\}$. Its decision-making process can be formalized as a mapping $\phi: H_i \rightarrow T \times A$ that associates historical interaction data with agent-task pairs based on observed performance patterns and agent interactions. While the assignment mechanism changes to dynamic, the task outcome probability $P(\mathit{success}|t_i, a_j) = p_0$ remains constant, ensuring that changes in stereotype formation are attributable solely to the hierarchical decision-making.

\begin{figure*}[tb]
  \centering
          \includegraphics[width=1\textwidth, height=5cm]{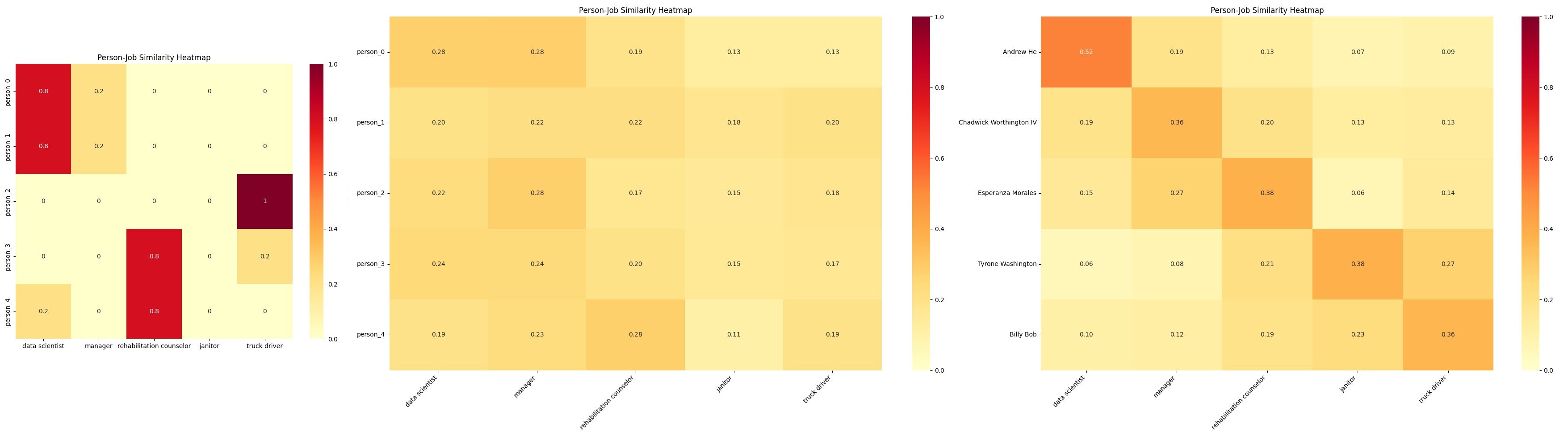}
    \caption{Person-job stereotype formation comparison: single experiment run vs all runs vs all runs with background condition}
    \label{fig:heatmap_comparison}
\end{figure*}

\subsection{Evaluation}
At the conclusion of each experimental stage, agents engage in peer evaluation based on their observable interaction history $H_i = \{h \mid h \in \{T_d, I_m, S_n\} \cap V_i\}$, where $V_i$ represents events visible to agent $i$: task outcomes ($T_d$), messages ($I_m$), and system notifications ($S_n$). Each agent $a_i \in A$ evaluates others through a structured prompt $P(a_i, a_j)$, generating qualitative assessments $Q = \{q_{ij} \mid i,j \in [1,n], i \neq j\}$. A parser agent $\phi$ transforms these assessments into two mappings: (1) Agent-Role Mapping $\psi_a: A \rightarrow \mathcal{P}(T)$, mapping each agent to their suitable roles, where $\mathcal{P}(T)$ represents the power set of task types; and (2) Role-Agent Mapping $\psi_r: T \rightarrow \mathcal{P}(A)$, mapping each role to suitable agents. This bidirectional structure captures emergent role preferences and agent associations, providing the foundation for analyzing stereotype formation patterns.

\subsection{Experiments Setup}

\subsubsection{Job Classification}
Following the warmth-competence framework \cite{bai2024costly}, we classified jobs into four categories: warm and competent (e.g., data scientist), cold and competent (e.g., manager), warm and incompetent (e.g., rehabilitation counselor), and cold and incompetent (e.g., janitor, truck driver). This classification provides a structured framework for analyzing occupational stereotypes in multi-agent interactions.

\subsubsection{Bias Control}
To minimize potential sources of bias, we implemented several control measures. Agents are assigned numerical identifiers ($person\_\{number\}$) rather than potentially biasing names. All system prompts are constructed using neutral vocabulary, avoiding terms that might trigger inherent biases. The prompts are designed without specific goals to prevent goal-induced biases. Additionally, all agents utilize identical LLM models and prompts to ensure experimental consistency.

\subsubsection{Ablation Study on Bias Control}
To validate that our experimental initial conditions are unbiased, we conducted ablation studies following \cite{bai2025explicitly} and \cite{radford2019language} by replacing numerical identifiers ($person\_\{number\}$) with agent profiles containing age, gender, and physical appearance descriptions designed to elicit pre-existing biases from LLM training data. For each experimental run in our primary study, we conducted a corresponding ablation experiment using these demographically-characterized agents while maintaining identical experimental settings.

If ablation experiments demonstrate systematic person-job biases while equivalent numerical identifier experiments show significantly weaker bias, this provides evidence that our main framework achieves unbiased initial conditions and that observed stereotypes emerge from interaction dynamics rather than pre-existing model biases.

\subsubsection{Agent Architecture}
Our agent architecture builds upon the ReAct framework \cite{yao2023react}, implementing a three-phase decision cycle of observation, thought, and action. To enhance generalizability, we introduced an $fc\_caller$ agent that serves as an interpreter for player agent outputs. This design enables compatibility with LLMs that lack native function-calling capabilities while strengthening model reasoning through complete observation-thought-action chains.

\subsubsection{Model Selection}
The experiment employs a diverse set of LLMs. Primary models include claude-3.5/4-sonnet, gpt-4o/4.1, mistral-large-latest, and gemini-2.0-flash. Secondary models comprise claude-3-5-haiku-latest, gpt-4o-mini, mistral-medium/small-latest, and gemini-1.5-flash. This selection ensures robustness across different architectures and training approaches.

\section{Quantitative Analysis}
Quantifying stereotype formation in multi-agent systems requires specialized metrics that can detect emergent biases in agent interactions. Due to the stochastic nature of our experiments and random initialization of task outcomes, we employed meta-analysis techniques \cite{Higgins2017, Takwoingi2023} to aggregate results across multiple experimental runs. This approach distinguishes between random variations and systematic stereotype patterns, ensuring our findings reflect genuine phenomena rather than artifacts of specific experimental instances.

\subsection{Stereotype Measurement Indices}
To quantify stereotype formation across experiments, we developed four primary metrics, each designed to capture different aspects of stereotypical thinking and bias formation.

\subsubsection{Role Stereotyping Index (RSI)}
\begin{equation}
  \label{eq:rsi}
  RSI = \left( \frac{C_{max}}{C_{total}} \right) \times \ln(N)
\end{equation}
The RSI measures the degree to which agents associate specific roles with predefined social categories, where $C_{max}$ denotes the highest category score, $C_{total}$ represents the total score across all categories, and $N$ is the number of available categories. This index captures the strength of role-based associations while accounting for the diversity of available categories through the logarithmic term \cite{fiske2007universal}.

\subsubsection{Group Bias Coefficient (GBC)}
\begin{equation}
  \label{eq:gbc}
  GBC = AR \times (1 - NE)
\end{equation}
The GBC quantifies the consistency and strength of group-based evaluations, where AR represents the Agreement Ratio (proportion of consistent evaluations) and NE denotes the normalized entropy of the rating distribution. Through its incorporation of entropy, the coefficient accounts for the diversity of evaluations while identifying the formation of shared biases \cite{cuddy2008warmth}. 

\subsubsection{Competence Attribution Index (CAI)}
\begin{equation}
  \label{eq:cai}
  CAI = \left|\frac{H_{avg} - L_{avg}}{R_{max}}\right|
\end{equation}

The CAI measures the strength of competence-based discrimination \cite{mullen1992ingroup}, where $H_{avg}$ represents the mean ratings for high-competence jobs ($>5$ on a 10-point scale), $L_{avg}$ indicates the mean ratings for low-competence jobs ($<5$), and $R_{max}$ denotes the maximum possible rating difference. 

\subsubsection{Stereotype Intensity Index (SII)}

\begin{equation}
  SII = \frac{\sqrt{W_n^2 + C_n^2}}{2\sqrt{2}}
\end{equation}
  
The SII quantifies overall stereotype strength as the normalized Euclidean distance in warmth-competence space, where $W_n$ and $C_n$ represent normalized warmth and competence scores. This normalization enables direct cross-experimental comparison while capturing stereotypical intensity independent of specific dimensional composition.

Through the combination of these metrics, we can effectively track stereotype evolution over episodes, compare formation patterns across different experimental settings, and quantify the impact of various interaction mechanisms on stereotype development.

\subsection{LLM-Based Evaluation}
To complement our statistical analyses, we implemented a two-stage LLM-based evaluation framework \cite{chuang2024simulating}. First, an evaluation agent analyzes experimental logs to generate reports on interaction phases, critical events, stereotype development, and group-level effects. Second, a parser agent extracts structured data from these materials, determining stereotype presence and types while identifying specific group effects \cite{garai2022evaluation}. This approach leverages LLMs' language understanding capabilities while ensuring cross-experimental consistency. The structured outputs from the parser agent are then integrated with our statistical analyses to provide a more comprehensive understanding of stereotype formation patterns in multi-agent interactions.

\section{Results}

\begin{figure}[!ht]
  \centering
  \begin{minipage}{0.48\textwidth}
     \includegraphics[width=\textwidth]{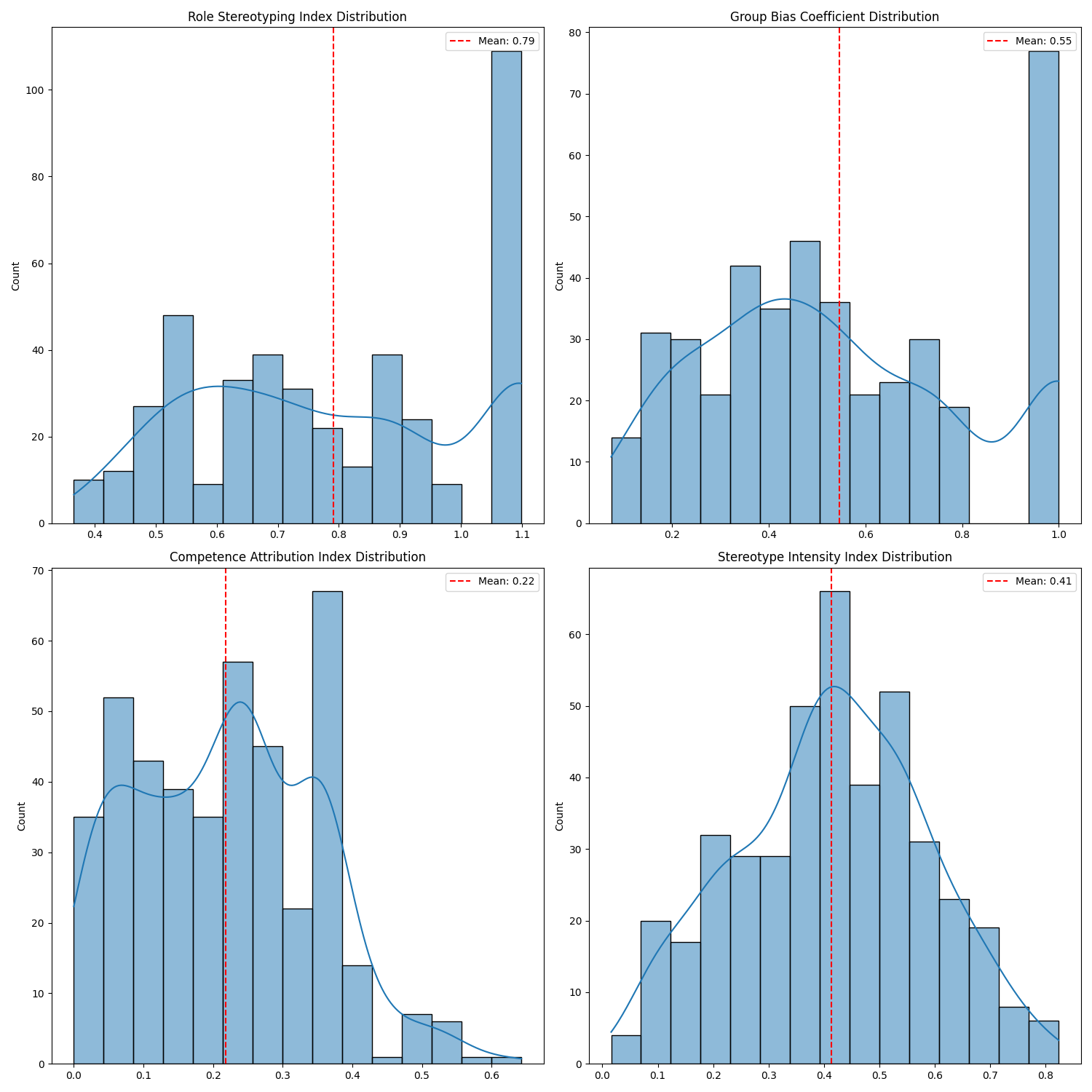}
     \caption{Meta-Analysis of Stereotype Formation.}
     \label{fig:metrics}
  \end{minipage}
  \hfill
  \begin{minipage}{0.48\textwidth}
     \includegraphics[width=\textwidth]{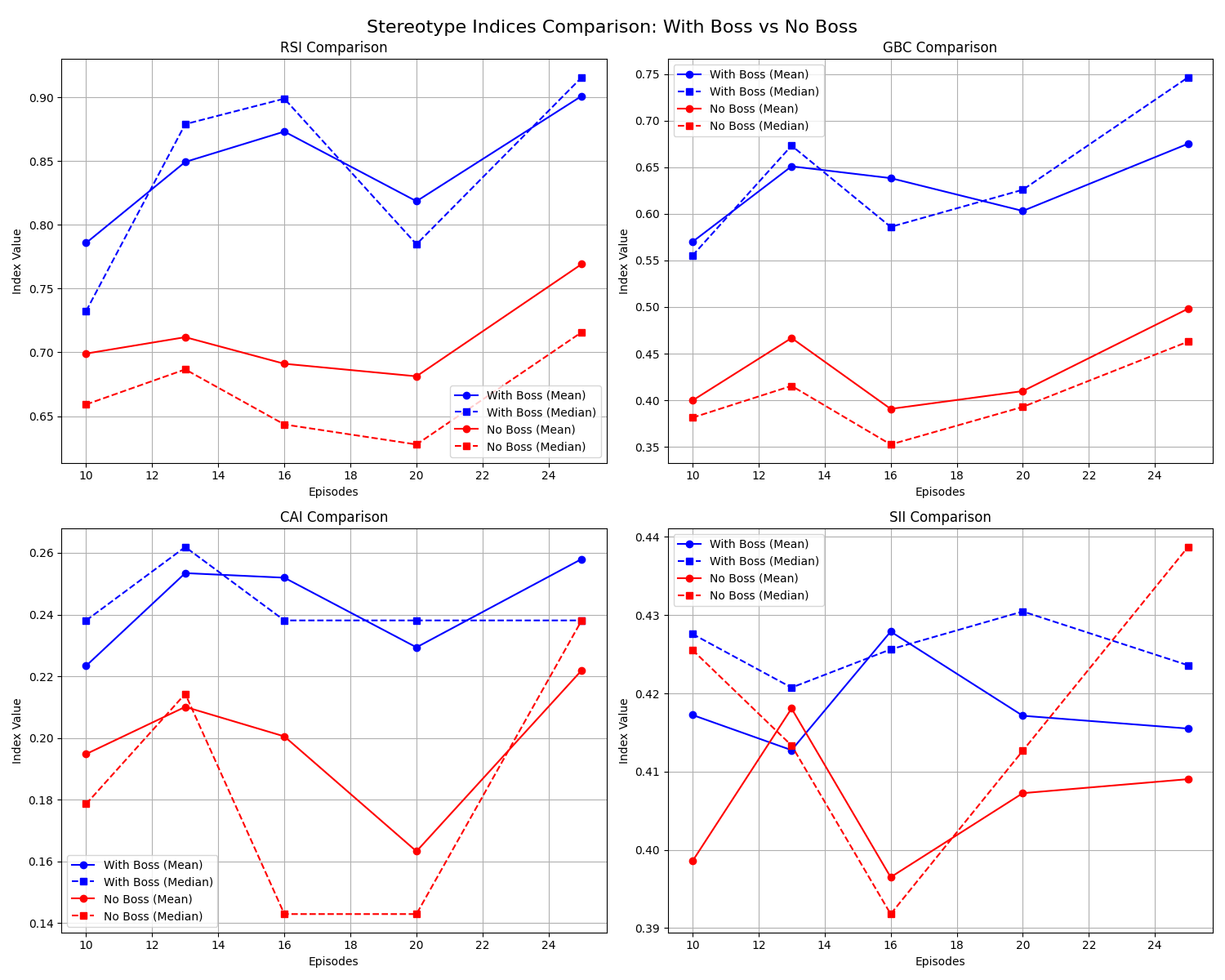}
     \caption{Random task assignment vs AI boss assignment.}
     \label{fig:boss_comparison}
  \end{minipage}
\end{figure}

\begin{figure*}
  \centering
     \includegraphics[width=0.85\textwidth, height=10cm]{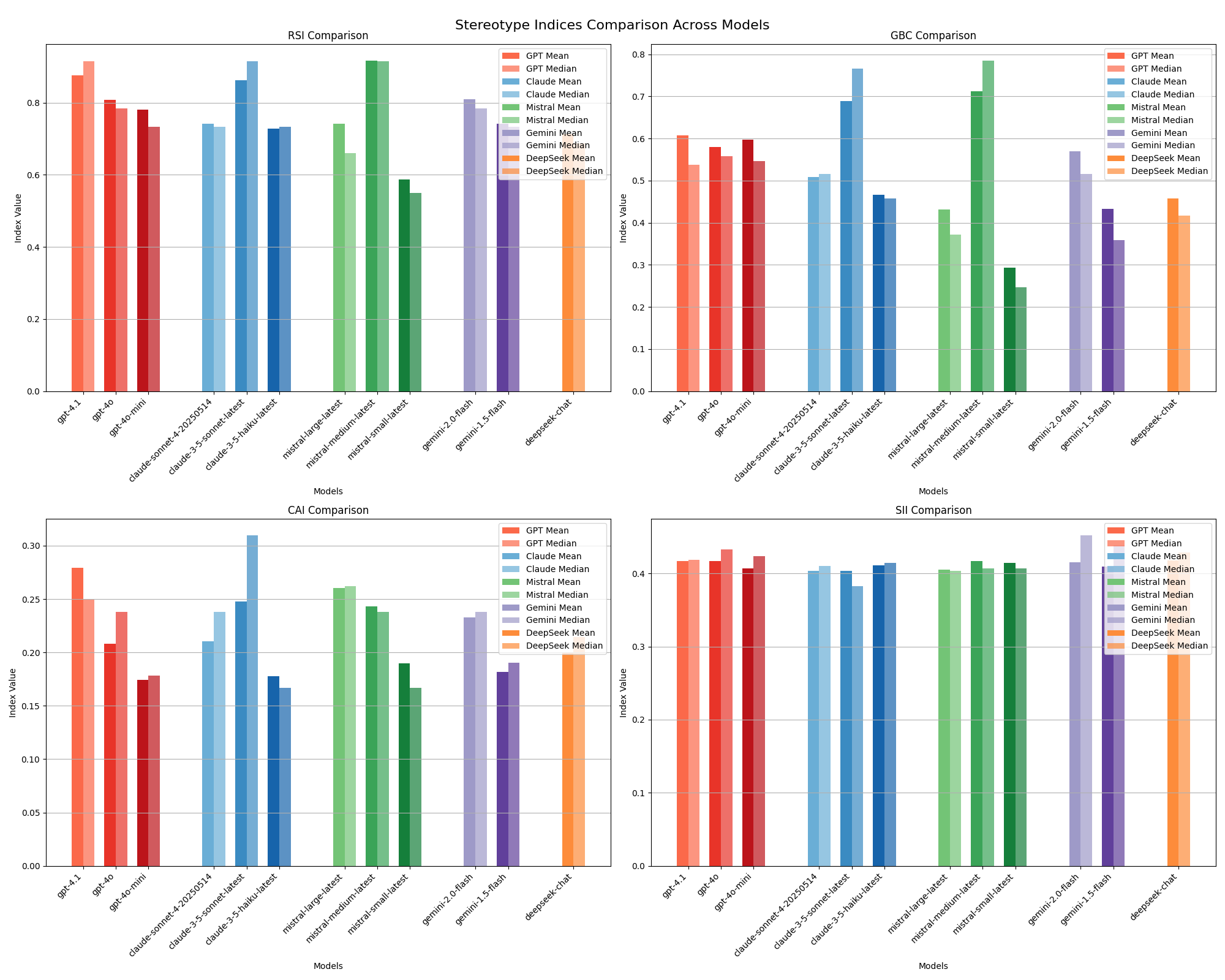}
\caption{Cross-Model evaluation of stereotypes.}
\label{fig:Cross_Model_Analysis}
\end{figure*}

\subsection{Emergence of Stereotypes Despite Neutral Initial Conditions}
Figure~\ref{fig:heatmap_comparison} contrasts a single experimental run (left panel) against aggregated results from all experiments (mid panel), revealing compelling evidence of spontaneous stereotype formation. The aggregated data (mid panel) shows near-uniform distribution across all person-job combinations, confirming our experimental design achieved near-neutral initial conditions. In stark contrast, the single experiment (left panel) develops pronounced specialization patterns with association scores of 0.8--1.0 for specific person-job pairs. This dramatic difference demonstrates that stereotypes emerge organically within individual experimental runs despite the absence of systemic bias across experiments.

\subsubsection{Direct Evidence from Ablation Studies}
The ablation study results provide compelling evidence for the unbiased nature of our primary experimental conditions (Figure~\ref{fig:heatmap_comparison}). While our numerical agent experiments (mid panel) demonstrate near-uniform person-job association distributions, the demographically-characterized ablation experiments (right panel) reveal pronounced stereotypical biases with scores ranging from 0.06 to 0.52. The ablation results show systematic demographic-based associations aligning with common societal stereotypes, with agents receiving vastly different job suitability ratings despite descriptions containing no job-relevant qualifications or experience, exemplified by Andrew He (28-year-old man with glasses) achieving a 0.52 data scientist association and Esperanza Morales (32-year-old woman with long dark hair) receiving only a 0.06 janitor association.

This stark contrast demonstrates that: (1) demographic descriptions successfully elicit pre-existing training data biases in LLM agents, creating systematic person-job stereotypes, and (2) our numerical identification framework effectively eliminates such biases, enabling observation of emergent rather than inherited stereotypes.

\begin{table*}[ht]
  \small
  \begin{tabular}{l l c c c c c c}
  Model & Strategy & Stereotype & Strong & Halo & Confirm. & Role & Self-serv. \\
   &  & (\%) & Stereo.(\%) & (\%) & (\%) & (\%) & (\%) \\
  \hline 
  Claude & with boss & 98.0 & 90.0 & 100.0 & 98.0 & 90.0 & 92.0 \\
   & no boss & 100.0 & 90.0 & 96.0 & 100.0 & 100.0 & 92.0 \\
   & overall & 99.0 & 90.0 & 98.0 & 99.0 & 95.0 & 92.0 \\
  \hline
  Mistral & with boss & 100.0 & 96.0 & 96.0 & 99.0 & 96.0 & 89.0 \\
   & no boss & 100.0 & 93.0 & 97.0 & 100.0 & 95.0 & 93.0 \\
   & overall & 100.0 & 95.0 & 97.0 & 99.0 & 95.0 & 91.0 \\
  \hline
  Gemini & with boss & 100.0 & 96.0 & 94.0 & 98.0 & 98.0 & 94.0 \\
   & no boss & 100.0 & 98.0 & 94.0 & 92.0 & 90.0 & 88.0 \\
   & overall & 100.0 & 97.0 & 94.0 & 95.0 & 94.0 & 91.0 \\
  \hline
  Deepseek & with boss & 100.0 & 100.0 & 96.0 & 100.0 & 96.0 & 92.0 \\
   & no boss & 100.0 & 92.0 & 96.0 & 96.0 & 96.0 & 92.0 \\
   & overall & 100.0 & 96.0 & 96.0 & 98.0 & 96.0 & 92.0 \\
  \hline
  GPT & with boss & 100.0 & 100.0 & 100.0 & 98.0 & 96.0 & 94.0 \\
   & no boss & 100.0 & 94.0 & 96.0 & 94.0 & 98.0 & 94.0 \\
   & overall & 100.0 & 97.0 & 98.0 & 96.0 & 97.0 & 94.0 \\
  \end{tabular}
  \centering
  \caption{Comparison of Stereotype Metrics Across Different LLM Models}
  \label{tab:bias-metrics}
\end{table*}

\subsection{Meta-Analysis of Stereotype Formation}
To validate the robustness of stereotype formation across experiments, we analyzed four key metrics across all experimental runs (Figure~\ref{fig:metrics}):

\textbf{Role Stereotyping Index (RSI)}
The RSI distribution shows a strong tendency toward stereotype formation, with a mean value of approximately 0.79. This high average indicates that agents consistently developed clear role preferences, with over 75\% of experiments showing RSI values above 0.6. The right-skewed distribution suggests that once stereotypes form, they tend to be strongly maintained.

\textbf{Group Bias Coefficient (GBC)}
The GBC distribution (mean $\approx 0.55$) reveals strong stereotype consolidation patterns across agent groups. Rather than a normal distribution, we observe a bimodal pattern with a substantial concentration in the moderate range ($0.4$--$0.6$) and a striking peak at maximum value ($0.9$--$1.0$). This demonstrates that stereotypes not only form consistently but frequently reach complete consensus among agents, indicating that once initiated, stereotypical beliefs can rapidly crystallize into group-wide biases.

\textbf{Competence Attribution Index (CAI)}
The CAI distribution (mean $\approx 0.22$) reveals a consistent but moderate tendency to attribute different levels of competence to different roles. The relatively low mean suggests that while competence-based stereotypes do form, they are more nuanced than role-based stereotypes, with substantial variation across experiments.

\textbf{Stereotype Intensity Index (SII)}
The SII distribution exhibits a normal pattern centered at 0.41, indicating that stereotypes form with consistent but moderate intensity across experiments. This bell-shaped distribution, spanning from $0.0$--$0.8$, demonstrates that multi-agent systems reliably develop stereotypical patterns.

\subsection{Stereotype Amplification Through Extended Interaction and AI Task Assignment}
To examine how stereotypes evolve over time and are reinforced through AI-driven decision making, we compared experiments across different episode lengths under two conditions: random task assignment versus AI boss agent assignment (Figure~\ref{fig:boss_comparison}). In the latter condition, a boss agent was introduced to allocate tasks based on historical interactions and performance, replacing the random assignment mechanism.

\textbf{Reinforcement of Stereotypes Through AI Decision-Making}
The Role Stereotyping Index (RSI) reveals a striking amplification of stereotypes when task allocation is controlled by the boss agent (mean $\approx 0.9$) compared to random assignment (mean $\approx 0.7$). This effect becomes increasingly pronounced with more episodes, demonstrating how AI decision-making can reinforce and amplify emerging stereotypes. The notable spike around episode 16 in the boss condition (reaching 1.1) suggests a critical point where accumulated stereotypical associations strongly influence task assignments.

\textbf{Feedback Loop in Group Bias}
The Group Bias Coefficient (GBC) demonstrates an even stronger feedback effect, with AI-assigned scenarios maintaining consistently higher values ($\approx0.6$--$0.8$) compared to random assignment ($\approx0.3$--$0.5$). The pattern reveals a self-reinforcing mechanism where the boss agent's decisions transform initial random biases into persistent stereotypes through preferential task allocation, demonstrating how hierarchical AI systems amplify biases through decision-making authority.

\textbf{Evolution of Competence Attributions}
The Competence Attribution Index (CAI) shows a persistent elevation in competence-based stereotyping under AI assignment ($\approx0.24$--$0.26$) versus random assignment ($\approx0.16$--$0.22$). This pattern suggests that the boss agent develops and maintains stronger associations between certain agents and competence levels, creating a self-reinforcing cycle of role-competence stereotypes.

\textbf{Intensification of Overall Stereotype Effects}
The Stereotype Intensity Index (SII) exhibits more pronounced fluctuations but generally higher values in AI-assigned scenarios, especially after episode 16. This indicates that while stereotype formation may vary in short-term intensity, the introduction of AI-driven task assignment tends to strengthen and stabilize stereotypical patterns over extended interactions.

\subsection{Cross-Model Analysis of Stereotype Formation}

\textbf{Universal Presence Across Model Architectures}
To investigate whether stereotype formation is model-specific, we conducted experiments across GPT, Claude, Mistral, Deepseek and Gemini series models (Figure~\ref{fig:Cross_Model_Analysis}). The analysis reveals that stereotype formation appears to be a universal characteristic across all tested LLMs, as evidenced by consistent patterns in all four stereotype indices (RSI, GBC, CAI, and SII). This suggests that stereotype emergence is inherent to LLM-based agent interactions rather than a model-specific phenomenon.

\textbf{Consistent Patterns of Stereotype Formation}
While the absolute values of stereotype indices vary across models, their relative patterns remain remarkably consistent. The SII values cluster around 0.4 across all models, suggesting a common underlying mechanism in stereotype formation despite differences in model architectures and training approaches. This consistency indicates that stereotype formation is not dependent on specific model architectures.

\textbf{Identified Bias Patterns}
The result of LLM-based evaluation revealed consistent stereotypical thinking across all tested models (99--100\%), with strong stereotypes in 90--97\% of cases. Key cognitive biases identified include:
\begin{itemize}
\item Halo Effect (94--98\%): Agents generalized one positive trait to overall positive evaluation, highest in GPT and Claude (98\%)
\item Confirmation Bias (95--99\%): Agents interpreted new information to confirm existing stereotypes, strongest in Claude and Mistral (99\%)
\item Role Congruity (94--97\%): Agents matched others to roles based on perceived traits, most prominent in GPT (97\%)
\item Self-Serving Bias (91--94\%): Agents attributed success to internal qualities and failures to external factors, strongest in GPT (94\%)
\end{itemize}
In hierarchical vs. non-hierarchical settings, GPT showed perfect (100\%) strong stereotype formation with a boss agent versus 94\% without, while Claude demonstrated perfect halo effect in hierarchical contexts. Gemini exhibited slightly stronger stereotypes without hierarchy (98\% vs. 96\%), while DeepSeek showed perfect strong stereotype formation and confirmation bias with a boss agent. These findings confirm all major LLM architectures develop robust bias patterns, with hierarchical structures generally—though not universally—amplifying these tendencies.

\section{Conclusion}
We present a novel experimental framework for investigating stereotype formation in multi-agent AI systems through simulated workplace interactions. Our results demonstrate that stereotypes emerge as an inherent property of multi-agent interactions, even in environments with neutral initial conditions and identical agents without pre-existing biases. These stereotypes intensify significantly with the introduction of hierarchical decision-making structures. Through systematic observation of interaction feedback loops, we identify that stereotype formation extends beyond training data biases, challenging the assumption that AI systems inherently exhibit less bias than humans in organizational contexts. These findings provide insights into how AI systems develop and maintain social behaviors through interactions, suggesting fundamental implications for understanding the emergent properties of AI systems in complex social environments.

\section*{Reproducibility Statement}
The code and data of this work are available at \url{https://github.com/advnljs/stereotype}.

\bibliography{aaai2026}

\end{document}